\documentclass[10pt,twocolumn,letterpaper]{article}

\usepackage{cvpr}
\usepackage{times}
\usepackage{epsfig}
\usepackage{graphicx}
\usepackage{amsmath}
\usepackage{amssymb}
\usepackage{algorithm}
\usepackage{algorithmic}
\usepackage{dsfont}
\usepackage{multirow}
\usepackage{enumitem}
\usepackage{cases}
\usepackage{bm}

\newcommand{\x}{\mathbf{x}}
\newcommand{\A}{\mathcal{A}}

\renewcommand{\S}{\mathcal{S}}

\newcommand{\rot}[1]{\rotatebox{90}{#1}}

\DeclareMathOperator{\st}{s.t.}
\DeclareMathOperator*{\argmax}{arg\,max}
\DeclareMathOperator*{\argmin}{arg\,min}

\usepackage[pagebackref=true,breaklinks=true,letterpaper=true,colorlinks,bookmarks=false]{hyperref}

\cvprfinalcopy 

\def\cvprPaperID{0872} 

\ifcvprfinal\fi
\begin{document}

\title{Uniform Information Segmentation}

\author{Radhakrishna Achanta, Pablo M{\'a}rquez-Neila,  Pascal Fua, and Sabine S\"{u}sstrunk \\
School of Computer and Communication Sciences (IC)\\
Ecole Polytechnique F\'{e}d\'{e}rale de Lausanne (EPFL)\\
CH-1015, Switzerland.\\
{\tt\small [radhakrishna.achanta,pablo.marquezneila,pascal.fua,sabine.susstrunk]@epfl.ch}}

\maketitle


\begin{abstract}

Size uniformity is one of the main criteria of superpixel methods. But size uniformity rarely conforms to the varying content of an image.  The chosen size of the superpixels therefore represents a compromise - how to obtain the fewest superpixels without losing too much important detail. We propose that a more appropriate criterion for creating image segments is information uniformity. We introduce a novel method for segmenting an image based on this criterion. Since information is a natural way of measuring image complexity, our proposed algorithm leads to image segments that are smaller and denser in areas of high complexity and larger in homogeneous regions, thus simplifying the image while preserving its details. Our algorithm is simple and requires just one input parameter - a threshold on the information content. On segmentation comparison benchmarks it proves to be superior to the state-of-the-art. In addition, our method is computationally very efficient, approaching real-time performance, and is easily extensible to three-dimensional image stacks and video volumes.

\end{abstract}

\section{Introduction}
Superpixels are a powerful preprocessing tool for image simplification. They reduce the number of image primitives from millions of pixels to a few thousands superpixels. Since their introduction~\cite{Ren_Malik_CVPR2003}, they have found their way into a wide-range of Computer Vision applications such as body model estimation~\cite{Mori_ModelSearch_ICCV2005}, multi-class segmentation~\cite{Gould_et_al_IJCV2008}, depth estimation~\cite{Zitnick_oversegmentation_IJCV2007}, object localization~\cite{Fulkerson_Vedaldi_Soatto_ICCV2009}, optical flow~\cite{MenzeGeiger_CVPR2015}, and tracking~\cite{Wang_etal_ICCV2011}. They provided an alternative to avoid the struggle with image semantics when using traditional segmentation algorithms~\cite{Comaniciu_Meer_PAMI2002,Felzenszwalb_Huttenlocher_IJCV2004}.

What differentiates superpixel algorithms from traditional segmentation algorithms is their ability to generate roughly equally-sized clusters of pixels. However, image-wide uniform size assumption obviously ignores the fact that real-world images do not have uniform visual complexity. Instead, they simultaneously feature highly variable, textured regions together with more homogeneous ones. As a consequence, superpixel methods over-segment texture-less areas and under-segment the textured regions. Thus, the price to pay for image-simplification using superpixels is that structures smaller than the chosen superpixel size have to be sacrificed.

In this paper we solve this problem with a simple alternative - information uniformity criterion. By creating clusters that contain roughly the same amount of information, we obtain image segments that are smaller in areas of high visual complexity and larger in areas of low visual complexity. Since our segments adapt to the image information content, we refer to
them as~\emph{adaptels}. Fig.~\ref{fig:teaser} shows an example segmentation. Note how the size of the adaptels conforms to the image content.

\newcommand{\teasercolwidth}{0.32\linewidth}

\begin{figure*}[htbp]
\centering
\begin{tabular}{c c c}
\includegraphics[width=\teasercolwidth]{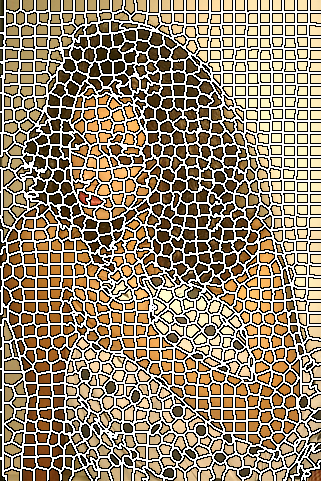}&
\includegraphics[width=\teasercolwidth]{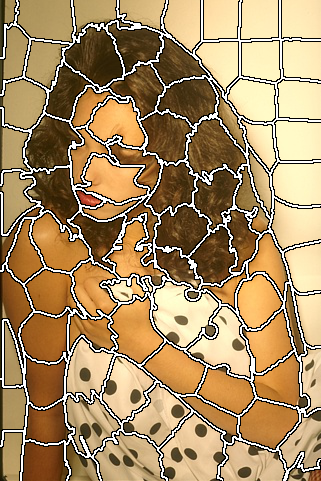}&
\includegraphics[width=\teasercolwidth]{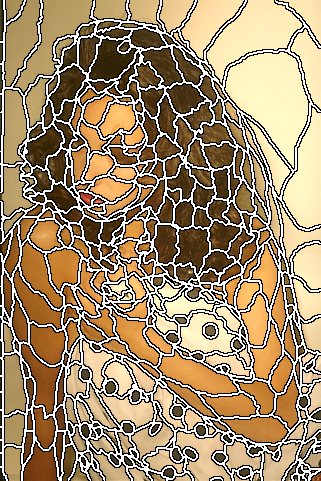}\\
(a) Too many superpixels & (b) Loss of detail & (c) Just right\\
\end{tabular}
\caption{The dilemma of choosing the right superpixel size - (a) retain detail and obtain too many superpixels, or (b) lose structures smaller than the superpixel size? Usually, several sizes are tried before choosing one that suits the need. (c) With adaptels, the choice of size is automatic, the number of superpixels is kept small, and all the advantages of a superpixel-based over-segmentation are retained. The number of segments in (c) is less than 50\% of those in (a).}
\label{fig:teaser}
\end{figure*}

Compared to the state-of-the-art, our method offers several advantages. There is no need to choose the size of the segments - their size is generated automatically. Similarly, the location and number of segments are automatically chosen to conform to the image content. Our algorithm minimizes the number of image segments while upper-bounding the information contained in them. This upper bound is the only parameter we need to set. Additionally, our algorithm grows segments ensuring connectivity. The resulting segments, adaptels, are compact, with a limited degree of adjacency. Finally, the algorithm complexity is linear in the number of pixels and runs in real-time without using any specialized hardware or optimization.

 The rest of the paper is as follows. Section~\ref{sec:sota} surveys the related background work. We present a formal explanation of adaptels and the corresponding algorithm in Section~\ref{sec:adaptels}. In Section~\ref{sec:comparison} we compare adaptels to other methods. We show that our method performs better than others on standard metrics. Section~\ref{sec:conclusion} concludes the paper.
\section{Related work}
\label{sec:sota}

Superpixel segmentation is an active research topic with large number of proposed methods. This review is not exhaustive but an attempt is being made to cover prominent algorithms that encompass a wide range of approaches for creating superpixels. We also include traditional segmentation algorithms that do not aim to create uniformly sized segments.

One of the earliest graph-based approaches, the Normalized cuts algorithm~\cite{Shi_Malik_NCuts_PAMI2000},
creates NCUTS superpixels by recursively computing normalized cuts for the pixel
graph. Felzenszwalb and Huttenlocher~\cite{Felzenszwalb_Huttenlocher_IJCV2004} propose a minimum spanning tree based EGB segmentation approach, which is computationally much simpler.
To create segments, which are essentially sub-trees, a stopping criterion is used to prevent the tree-growing from spanning the entire image with a single tree. Unlike NCUTS, EGB does not create uniformly-sized superpixels.  Moore et al.~\cite{Moore_Prince_Warrel_Mohammad_Jones_CVPR2008}
generate SLAT superpixels by finding the shortest paths that split the image
into vertical and horizontal
strips. Similarly, Zhang et
al.~\cite{Zhang_etal_ICCV2011} create SPBO superpixels by applying horizontal
and vertical graph-cuts to overlapping strips of an image. Instead of finding
cuts on an image, Veskler et al.~\cite{Veksler_Superpixels_ECCV2010},  generate
GCUT superpixels by stitching together overlapping image patches using graph cuts optimization. More recently, Liu
et al.~\cite{Liu_etal_ERS_CVPR2011} present another graph-based approach to create ERS superpixels
that connects subgraphs by maximising the entropy rate of a random walk.

There are several other algorithms that are not graph-based. The watershed
algorithm~\cite{Vincent_Soille_PAMI1991} accumulates similar pixels
starting from local minima to find WSHED segments. The mean shift algorithm~\cite{Comaniciu_Meer_PAMI2002}
iteratively locates local maxima of a density function in color and image plane
space. Pixels that lead to the same local maximum belong to the same MSHIFT segment.
Quick-shift~\cite{Vedaldi_Soatto_QuickShift_ECCV2008} creates QSHIFT
superpixels by seeking local maxima like MSHIFT but is more efficient in terms
of computation. The Turbopixels algorithm~\cite{Levenshtein_Kutulakos_Fleet_Dickinson_Siddiqi_PAMI2009}
generates TPIX superpixels by progressively dilating pixel seeds located at regular
grid centers using a level-set approach. Like TPIX, the Simple Linear Iterative
Clustering (SLIC) algorithm~\cite{Achanta_etal_SLIC_PAMI2012} also relies on
starting seed pixels chosen at regular grid intervals. It performs a localized
k-means optimization in the five-dimensional CIELAB color and image space to
cluster pixels into SLIC superpixels. Two recent variants of SLIC are presented by Li and Chen~\cite{Li_Chen_CVPR2015}, and by Liu et al.~\cite{Liu_etal_CVPR2016}. The former projects the five-dimensional space of spatial coordinates and color on a ten-dimensional space and the latter projects it to a two-dimensional space before performing $k$-means clustering. Both the works show improvement in segmentation quality.

\newcommand{\gapx}{@{\hspace{3.5mm}}}
\begin{table*}
\begin{tabular}{\gapx r \gapx c \gapx c \gapx c \gapx c \gapx c \gapx c \gapx c \gapx c \gapx c \gapx c \gapx c \gapx c \gapx c \gapx c \gapx c \gapx c}
\\
\hline
\rot{Method}  & \rot{EGB} & \rot{MSHIFT} & \rot{WSHED} & \rot{QSHIFT} & \rot{GEOD} & \rot{NCUTS} & \rot{SLAT} & \rot{SPBO} & \rot{GCUT} & \rot{TPIX} & \rot{SLIC} & \rot{SEEDS} & \rot{ERS} & \rot{LSC} & \rot{MSLIC} & \rot{\textbf{Adaptels}} \\
Reference & \cite{Felzenszwalb_Huttenlocher_IJCV2004} & \cite{Comaniciu_Meer_PAMI2002} & \cite{Vincent_Soille_PAMI1991} & ~\cite{Vedaldi_Soatto_QuickShift_ECCV2008} & ~\cite{Wang_etal_StructureSensitive_IJCV2013} & \cite{Shi_Malik_NCuts_PAMI2000} & \cite{Moore_Prince_Warrel_Mohammad_Jones_CVPR2008} & \cite{Zhang_etal_ICCV2011} & \cite{Veksler_Superpixels_ECCV2010} & \cite{Levenshtein_Kutulakos_Fleet_Dickinson_Siddiqi_PAMI2009} & \cite{Achanta_etal_SLIC_PAMI2012} & \cite{Bergh_etal_SEEDS_IJCV2015} & \cite{Liu_etal_ERS_CVPR2011} & \cite{Li_Chen_CVPR2015} & \cite{Liu_etal_CVPR2016} & $-$ \\
\hline
\\
Agglomerative & $\bullet$ & $\bullet$ & $\bullet$ & $\bullet$ & $\bullet$ & $\cdot$ & $\cdot$ & $\cdot$& $\cdot$& $\bullet$& $\cdot$& $\cdot$& $\cdot$ & $\cdot$ & $\cdot$ & $\bullet$\\
Divisive & $\cdot$ & $\cdot$ & $\cdot$ & $\cdot$ & $\bullet$ & $\bullet$ & $\bullet$ & $\bullet$ & $\cdot$ & $\cdot$ & $\cdot$ & $\bullet$ & $\cdot$ & $\cdot$ & $\cdot$ & $\cdot$\\
Graph-based & $\bullet$ & $\cdot$& $\cdot$& $\cdot$& $\cdot$& $\bullet$ & $\cdot$& $\bullet$& $\bullet$& $\cdot$& $\cdot$& $\cdot$& $\bullet$ & $\cdot$ & $\cdot$ & $\cdot$\\
Patch-based & $\cdot$ & $\cdot$ & $\cdot$ & $\cdot$ & $\cdot$ & $\cdot$ & $\cdot$ & $\cdot$ & $\bullet$ & $\cdot$& $\cdot$& $\cdot$& $\cdot$ & $\cdot$ & $\cdot$ & $\cdot$\\
Center-seeking & $\cdot$& $\bullet$ & $\cdot$& $\bullet$ & $\bullet$ & $\cdot$& $\cdot$& $\cdot$& $\cdot$& $\cdot$& $\bullet$ & $\cdot$& $\cdot$& $\bullet$ & $\bullet$ & $\cdot$\\
Border seeking & $\cdot$& $\cdot$& $\bullet$ & $\cdot$& $\cdot$& $\cdot$& $\cdot$& $\cdot$& $\cdot$& $\bullet$ & $\cdot$& $\cdot$& $\cdot$& $\cdot$ & $\cdot$ & $\bullet$\\
Iterative & $\cdot$& $\cdot$& $\cdot$& $\cdot$& $\bullet$& $\cdot$& $\cdot$& $\cdot$& $\cdot$& $\cdot$& $\bullet$ & $\bullet$ & $\cdot$ & $\bullet$ & $\bullet$ & $\cdot$ \\
Grid seeding & $\cdot$& $\cdot$& $\cdot$& $\cdot$& $\bullet$ & $\cdot$& $\bullet$ & $\bullet$ & $\bullet$ & $\bullet$ & $\bullet$ & $\bullet$ & $\cdot$& $\bullet$ & $\bullet$ & $\cdot$ \\
Uniform size & $\cdot$ & $\cdot$& $\cdot$ & $\cdot$& $\cdot$& $\bullet$& $\bullet$& $\bullet$& $\bullet$& $\bullet$& $\bullet$ & $\bullet$ & $\cdot$& $\bullet$ & $\bullet$ & $\cdot$\\
Real-time & $\bullet$ & $\cdot$& $\bullet$ & $\cdot$& $\cdot$& $\cdot$& $\cdot$& $\cdot$& $\cdot$& $\cdot$& $\bullet$ & $\bullet$ & $\cdot$& $\cdot$ & $\bullet$ & $\bullet$ \\
\\
\hline
\\
\end{tabular}
\caption{A comparison by algorithmic properties of the different superpixel
methods considered in this review. A large dot indicates the presence of the corresponding characteristic.}
\label{table:sota_comparison}
\end{table*}

Wang et al.~\cite{Wang_etal_StructureSensitive_IJCV2013}
present a geodesic distance based algorithm that generates GEOD superpixels of varying
size based on image content but is slow in practice. A more recent non-graph algorithm~\cite{Bergh_etal_SEEDS_IJCV2015} generates SEEDS superpixels by iteratively
improving an initial rectangular approximation of superpixels using coarse to
fine pixel  exchanges  with neighboring superpixels.

A comparative summary of the state-of-the-art is presented in Table~\ref{table:sota_comparison}. EGB, MSHIFT, and WSHED are traditional segmentation algorithms that do not aim
for uniformly-sized, compact segments. Of the others, NCUTS, SLAT, TPIX, SLIC, and SPBO are more compact. SLIC,
ERS, and SEEDS perform well on benchmark comparisons. EGB, SLIC, and SEEDS are
the fastest in computation. TPIX, SLIC, ERS, and SEEDS allow the user control
over the number of output segments.  This last property of superpixels is important because it lets the user choose the size of the superpixels
based on needs of the application. By doing so, the user accepts to lose
structural information finer than the superpixel size. The Adaptel algorithm
is the only one we are aware of that frees the user from making
this choice, and yet offers compactness, high precision, and computational
efficiency.


\section{The proposed method}
\label{sec:adaptels}
Image segmentation is the procedure of grouping the pixels~$\mathcal{P}$ of
an image into a number of connected components~$\{\mathcal{A}_k\}_{k=1}^{K}$
that form a partition.
Our goal is to minimize the number of segments~$K$ 
while bounding the amount of information each one contains. We therefore want
to find

\begin{subnumcases}{\label{eq:theproblem} \min_{\{\A_k\}_{k=1}^{K}} \!\!\! K \!\!\!\!\!\!\!\!\! \qquad \st}
\label{eq:theproblem_partition}
\bigcup_k \A_k\!\! = \!\!\mathcal{P}, \!\!\!\!\! \quad \A_j \! \cap \! A_k \!\! = \!\! \emptyset\quad \!\! \forall j \!\!\neq \!\!k, \\
\label{eq:theproblem_connected}
\mathrm{Connected}(\A_k) = 1 \ \forall k,\\
\label{eq:theproblem_info}
I(\A_k) \le T \ \forall k.
\end{subnumcases}

The resulting~$\A_k$ are the adaptels.
Line~\eqref{eq:theproblem_partition} ensures that they form an image-partition.
Line~\eqref{eq:theproblem_connected} enforces connectivity of
every adaptel. Line~\eqref{eq:theproblem_info} constraints the amount of information contained
in every adaptel to be less than a threshold~$T$, given in bits.
The value of $T$~is the only hyperparameter of our method. It
controls the number of adaptels in the resulting segmentations.

We define the information contained in an adaptel to be the Shannon \emph{self-information}
\begin{equation}
    \label{eq:information}
    I(\A_k) = \min_{\theta} -\log P(\A_k; \theta), 
\end{equation}
for a given family of probability distributions~$P$ parameterized by a
vector~$\theta$. $I(\A_k)$~measures the complexity of an adaptel~$\A_k$ by finding
the distribution~$P(\cdot; \theta)$ from the family~$P$ that best fits its content and then
computing the number of bits required to encode $\A_k$ under that distribution.
Instead of imposing a fixed
criterion on complexity, our method is flexible enough to allow the user to define
what complexity means by choosing the proper family of distributions~$P$:
simple regions are those that are likely to happen under~$P$, while complex
regions are those with low probability.
To illustrate, a Gaussian distribution over the color space will consider homogeneous-looking regions to 
be \emph{simple}, while for a bimodal distribution \emph{simple} means
regions made up of two colors.
By thresholding the information content, we make adaptels small
in areas of large complexity and vice versa.

The joint probability~$P(\A_k; \theta)$ depends on tens or hundreds of variables,
which makes its exact representation and computation intractable. We will thus make the
usual i.i.d.~assumption over pixels inside each adaptel and factorize
the joint probability as
\begin{equation}
    \label{eq:information2}
    I(\A_k) = \min_{\theta} -\sum_{i\in\A_k} \log P(\x_i; \theta),
\end{equation}
where the feature vector~$\x_i$ for pixel~$i$ contains a description
of its visual appearance and position.


\subsection{Adaptel Algorithm}
The minimization problem of Eq.~\eqref{eq:theproblem} is NP-hard and intractable even
for small images. We approach the
problem in an approximate manner and grow the adaptels as shown in
Algorithm~\ref{algo:adaptel}.

Every adaptel~$\A_k$ starts growing from a single pixel called its \emph{seed}, which is used to initialize a candidate map~$C$. At every iteration we take the pixel~$c$ from the candidate map~$C$ that contributes the least information to the adaptel. We add $c$ to the adaptel and update the map~$C$ with the neighbor pixels of~$c$ as possible candidates. The adaptel stops growing when the information threshold~$T$ is reached. This growing procedure ensures the conditions of connectivity~\eqref{eq:theproblem_connected} and bounded
information~\eqref{eq:theproblem_info}.


To mitigate the greediness of our algorithm,
we allow neighboring adaptels to compete with each other for pixel ownership.
This helps ensure that pixels are assigned to the most appropriate adaptel
and not to the one that reaches them first. To this end, we keep a map ~$D$ that stores the
accumulated information contribution of each pixel when it joins an adaptel (line~\ref{algo:mapD} in Algo.~\ref{algo:adaptel}). Candidate pixels are
reassigned to the current adaptel if their new information contribution
is lower than the one stored in the map~$D$ (line~\ref{algo:reassign} in Algo.~\ref{algo:adaptel}).
As an additional consequence,
the information map~$D$ also encourages the compactness
of the adaptels.

\begin{algorithm}[H]
\caption{{\sc GrowAdaptel}: Growing an adaptel~$\A_k$.}
\begin{algorithmic}[1]
\REQUIRE The upper bound~$T$, the adaptel seed pixel~$s$, the information map~$D$, connected neighbors $\mathcal{N}$ for each pixel
\ENSURE The adaptel~$\A_k$, updated information map~$D$
\STATE Initialize adaptel $\A_k \gets \emptyset$
\STATE Initialize candidate map $C[s] = I\left(\{s\}\right)$ \label{algo:adaptel:initinfo}
\WHILE {$C$ is not empty}
    \STATE Get best candidate $c \gets \argmin_c C[c]$
    \STATE Put candidate in adaptel $\A_k \gets \A_k \cup \{c\}$ \label{algo:adaptel:append}
    \STATE Update information map $D[c] \gets C[c]$
    \label{algo:mapD}
    \FOR {$c' \in \mathcal{N}(c)$}
        \STATE Compute information of the adaptel with the new candidate $e \gets I\left(\A_k \cup \{c'\}\right)$ \label{algo:adaptel:info}
        \STATE Update candidate map $C[c'] \gets e$ \textbf{if} $e < T$ \AND $e < D[c']$
        \label{algo:reassign}
    \ENDFOR
    \STATE Remove $c$ from $C$
\ENDWHILE
\RETURN $\A_k, D$
\end{algorithmic}
\label{algo:adaptel}
\end{algorithm}

Algorithm~\ref{algo:adaptels} repeats this procedure as many times as necessary until
the whole image is partitioned, thus meeting the constraint
of Eq.~\eqref{eq:theproblem_partition}.
The seed of every adaptel is chosen among
the border pixels of the previous ones, stored in the
list~$\mathcal{S}$. We set the seed of the first adaptel to be the
central pixel of the image.
Fig.~\ref{fig:adaptel_growing} illustrates the complete procedure.

\begin{algorithm}[H]
\caption{The Adaptel algorithm.}
\begin{algorithmic}[1]
\REQUIRE Information upper bound~$T$, seed pixel~$s$
\ENSURE Set of adaptels~$\aleph$
\STATE $\aleph \gets \emptyset$
\STATE Set of seeds $\S \gets \{s\}$
\STATE Initialize information map $D[:] \gets \infty$
\WHILE {$\S$ is not empty}
\STATE Get seed $s \in \S$
\STATE $\A_k, D \gets \textsc{GrowAdaptel}(T, s, D)$
\STATE $\aleph \gets \aleph \cup \{\A_k\}$
\STATE $\S \gets \S \cup \left\{\textrm{Pixels bordering } \A_k\right\}$ \label{algo:adaptels:border}
\STATE Remove seeds from $\S$ that are assigned to an adaptel $\S \gets \S - \bigcup_{\A_k\in\aleph}\A_k$ \label{algo:adaptels:rmseeds}
\ENDWHILE
\RETURN$\aleph$
\end{algorithmic}
\label{algo:adaptels}
\end{algorithm}

\newcommand{\ggap}{@{\hspace{.5mm}}}
\newcommand{\expcolwidth}{0.245\linewidth}
\begin{figure*}[htbp]
\centering
\begin{tabular}{\ggap c \ggap c \ggap c \ggap c}
\includegraphics[width=\expcolwidth]{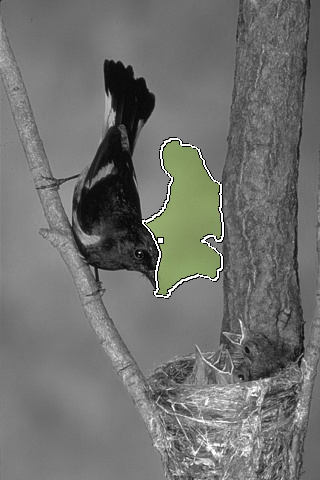}&
\includegraphics[width=\expcolwidth]{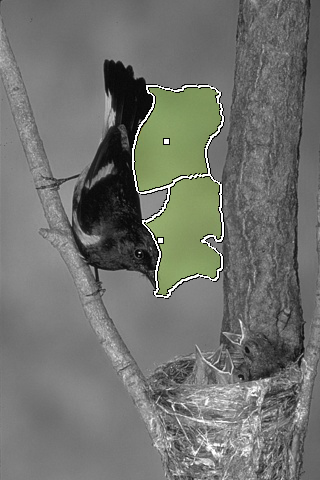}&
\includegraphics[width=\expcolwidth]{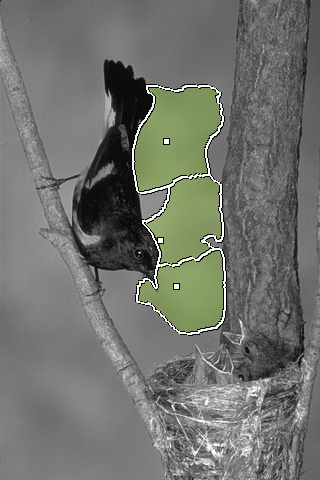}&
\includegraphics[width=\expcolwidth]{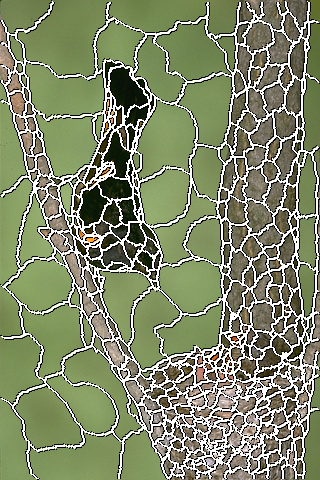}\\
(a) & (b) & (c) & (d)\\
\end{tabular}
\caption{The process of segmenting an image. (a) The first adaptel is grown from the center seed (white square) outwards by gathering neighboring pixels in order of increasing information until the information upper-bound $T$ is reached. (b and c) Subsequent adaptels are grown using pixels at the borders of the previous adaptels as seeds (white squares). As seen, the new adaptels capture pixels from the previous ones, resulting in constantly evolving segment boundaries. This mitigates the greediness of the algorithm and encourages boundary adherence. (d) Final segmented image. }
\label{fig:adaptel_growing}
\end{figure*}
By growing the adaptels using the least informative candidate until they reach the information threshold,
we are maximizing their sizes. In turn, this minimizes the number of adaptels,
as expressed in the initial formulation of our problem (Eq.~\eqref{eq:theproblem}).

Computing the information in lines~\ref{algo:adaptel:initinfo}
and~\ref{algo:adaptel:info} of Algorithm~\ref{algo:adaptel}
depends on the specific family of probability distributions chosen.
In the next section we introduce the family of probability distributions
we use in our experiments, and describe how to compute the information
of growing adaptels with it.


\subsection{Accommodating the visual complexity of natural images}

The Adaptel algorithm is agnostic to the family of probability
distributions~$P$ used to describe complexity. Different families lead to
different styles of superpixels. However, looking for the suitable distribution
for every particular case is hard. In this section we introduce a family of
probability distributions that are suitable for a
wide range of applications.

In most real-world cases, complexity is perceived as visual variability.
Homogeneous, constant-color regions are perceived as simple, while textured
regions are complex. Based on this idea, we look for a unimodal family of
distributions in the space of visual features.

The most straightforward family of distributions in this category are the
multivariate normal distributions. However, the normal
distribution over-penalizes pixels that are visually far from the mean, while it
under-penalizes pixels close to the mean. This leads to extreme segmentations
with very large adaptels in homogeneous areas and very small ones in
textured areas.

A unimodal family with smoother behavior
is the multivariate double exponential distribution
\begin{equation}
    \label{eq:neila_distribution}
    P_{mde}(\x; \bm{\mu}, \sigma) = \frac{1}{Z} \exp\left(-{\sqrt{\dfrac{(\x - \bm{\mu})^T(\x - \bm{\mu)}}{\sigma^2}}}\right),
\end{equation}
where $Z$~is the normalization factor and~$\x$ is the vector of visual features.
Unlike the normal, the double exponential has a
softer decay when moving away from the mean, and it does not penalize variability
as much. Using it yields better-behaved
segmentations, with more regularity on the sizes of the adaptels. This will be our distribution of choice for the remainder of this paper.

The double exponential distribution is parameterized by the mean vector~$\bm{\mu}$ and
the scalar variance~$\sigma^2$. We fix the variance~$\sigma^2$ so that~$P_{mde}(\bm{\mu}; \bm{\mu}, \sigma)=1$,
leaving remaining vector~$\bm{\mu}$ as the only parameter~$\theta$ of our family.
Computing the information of an adaptel with Eq.~\eqref{eq:information2}
requires finding the parameter~$\theta$ that best fits the
content of the adaptel. With the
multivariate double exponential, the best fit is given by the mean of the observations
inside the adaptel,
\begin{equation}
    \theta^*(\A_k) = \dfrac{1}{|\A_k|} \sum_{i\in\A_k} \x_i,
\end{equation}
and the information is
\begin{equation}
    I(\A_k) = -\log \sum_{i\in\A_k} P_{mde}(\x_k; \theta^*(\A_k)).
\end{equation}
These two equations give the implementation of lines~\ref{algo:adaptel:initinfo}
and~\ref{algo:adaptel:info} of Algorithm~\ref{algo:adaptel}. For efficiency,
we do not compute the information from scratch at
every iteration in line~\ref{algo:adaptel:info},
but perform online updates starting from previous estimations of the
mean and the information.

All the results shown in this paper are obtained with this family of double exponential
distributions with fixed~$\sigma$. Reasonable values of the upper-bound~$T$ for
natural images with this family are in range $(50, 150)$~bits. Source code will be made available upon publication of the paper.

\newcommand{\viscompwd}{0.108\linewidth}
\newcommand{\gap}{@{\hspace{0.5mm}}}
\begin{figure*}
\centering
\begin{tabular}{\gap r \gap r \gap r \gap r \gap r \gap r \gap r \gap r \gap r}

\includegraphics[width=\viscompwd]{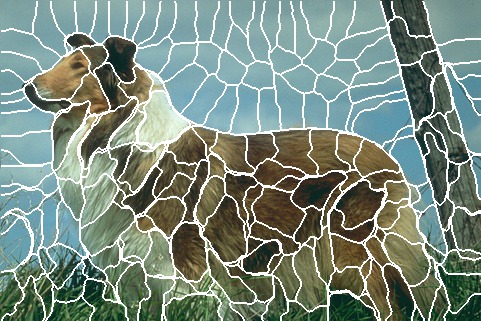}&
\includegraphics[width=\viscompwd]{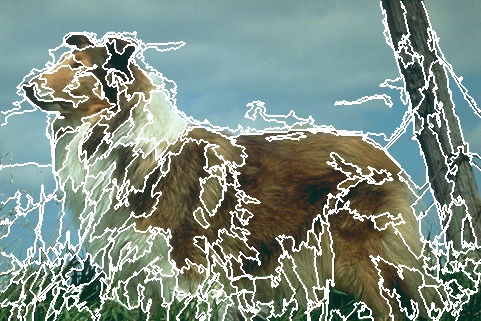}&
\includegraphics[width=\viscompwd]{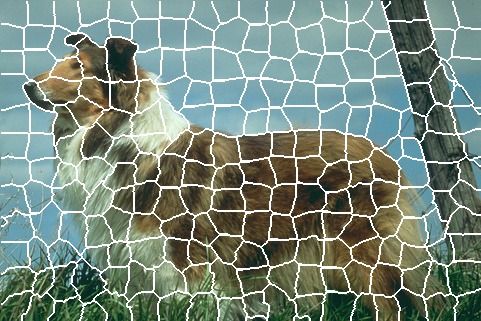}&
\includegraphics[width=\viscompwd]{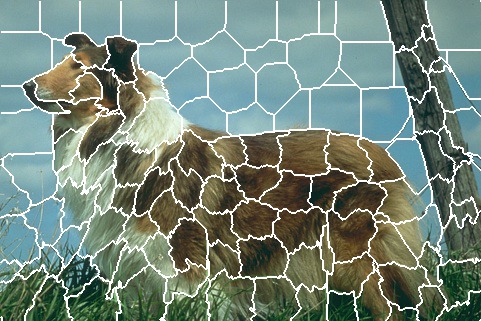}&
\includegraphics[width=\viscompwd]{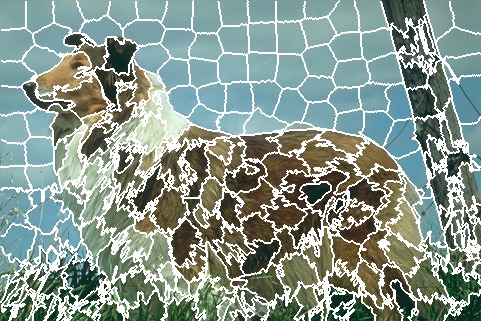}&
\includegraphics[width=\viscompwd]{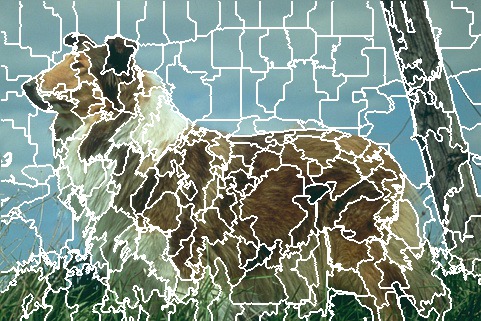}&
\includegraphics[width=\viscompwd]{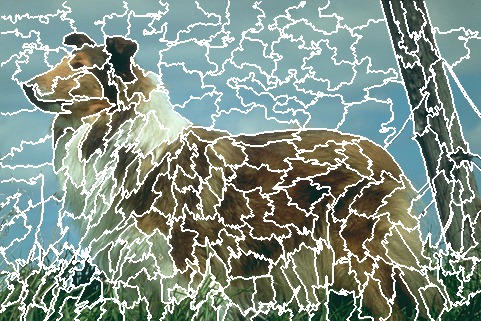}&
\includegraphics[width=\viscompwd]{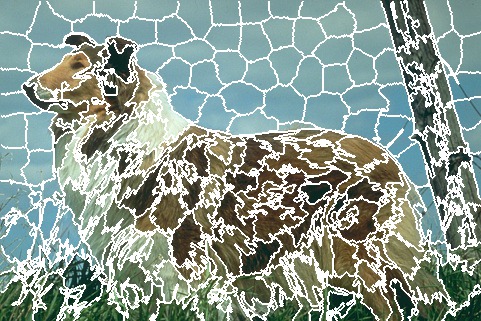}&
\includegraphics[width=\viscompwd]{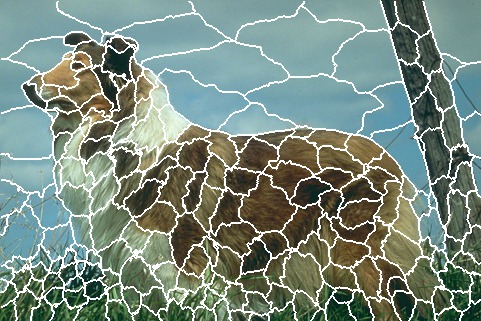}\\

\includegraphics[width=\viscompwd]{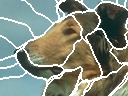}&
\includegraphics[width=\viscompwd]{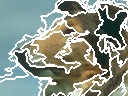}&
\includegraphics[width=\viscompwd]{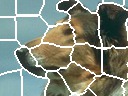}&
\includegraphics[width=\viscompwd]{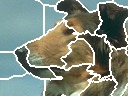}&
\includegraphics[width=\viscompwd]{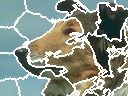}&
\includegraphics[width=\viscompwd]{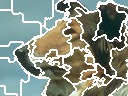}&
\includegraphics[width=\viscompwd]{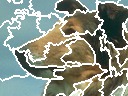}&
\includegraphics[width=\viscompwd]{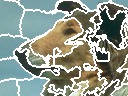}&
\includegraphics[width=\viscompwd]{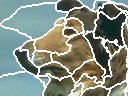}\\

\includegraphics[width=\viscompwd]{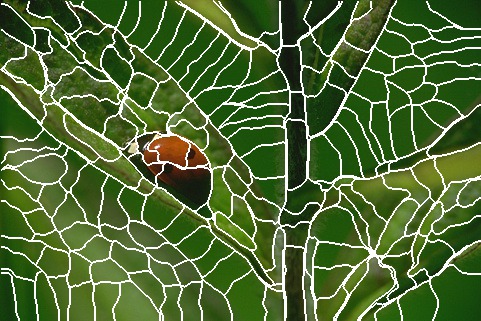}&
\includegraphics[width=\viscompwd]{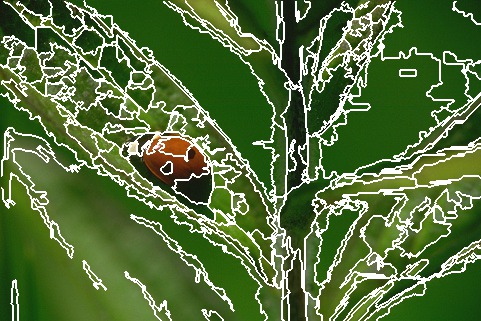}&
\includegraphics[width=\viscompwd]{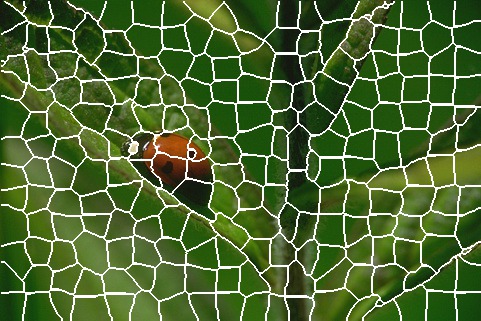}&
\includegraphics[width=\viscompwd]{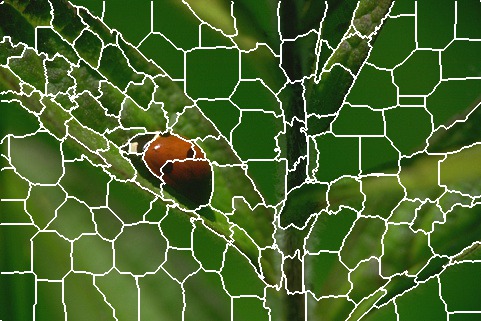}&
\includegraphics[width=\viscompwd]{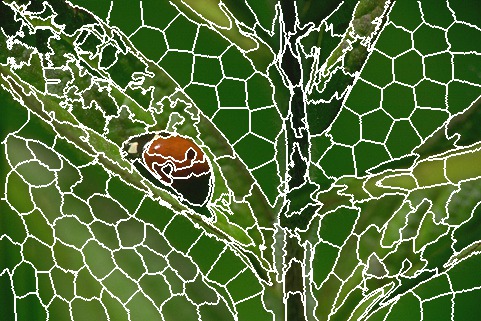}&
\includegraphics[width=\viscompwd]{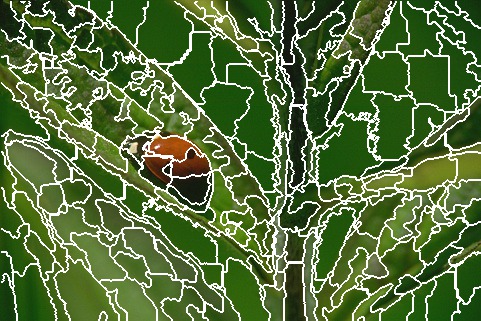}&
\includegraphics[width=\viscompwd]{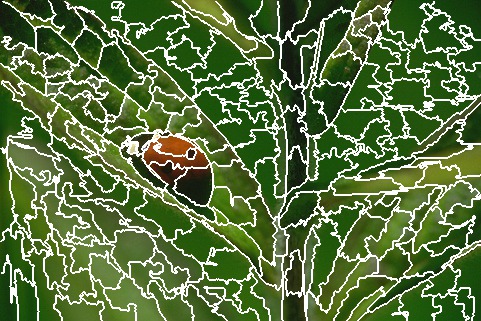}&
\includegraphics[width=\viscompwd]{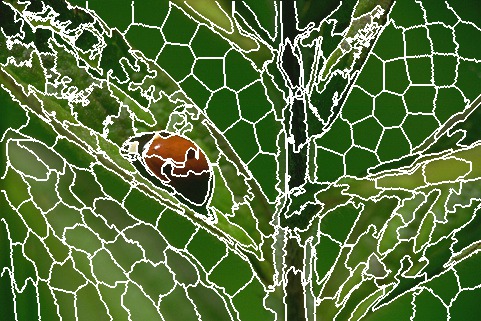}&
\includegraphics[width=\viscompwd]{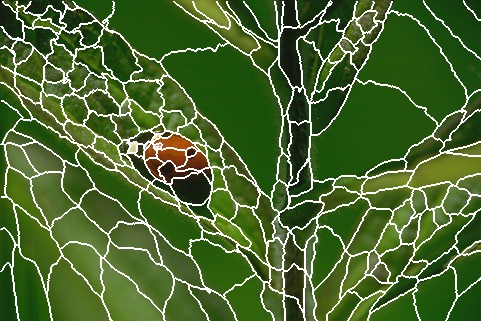}\\

\includegraphics[width=\viscompwd]{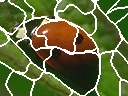}&
\includegraphics[width=\viscompwd]{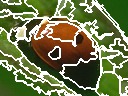}&
\includegraphics[width=\viscompwd]{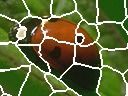}&
\includegraphics[width=\viscompwd]{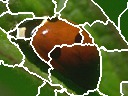}&
\includegraphics[width=\viscompwd]{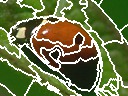}&
\includegraphics[width=\viscompwd]{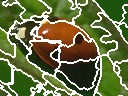}&
\includegraphics[width=\viscompwd]{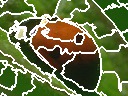}&
\includegraphics[width=\viscompwd]{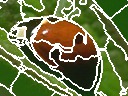}&
\includegraphics[width=\viscompwd]{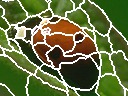}\\

\includegraphics[width=\viscompwd]{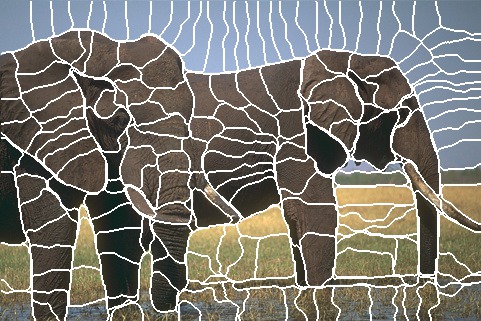}&
\includegraphics[width=\viscompwd]{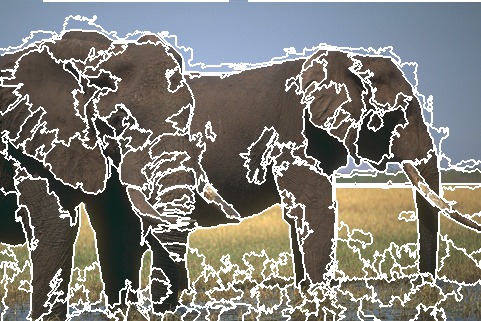}&
\includegraphics[width=\viscompwd]{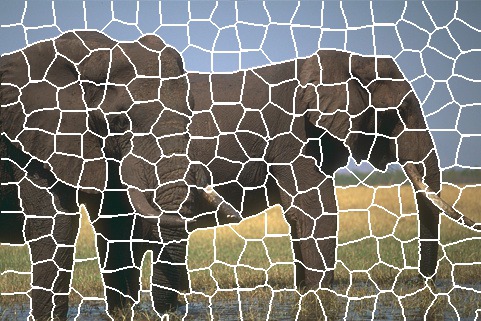}&
\includegraphics[width=\viscompwd]{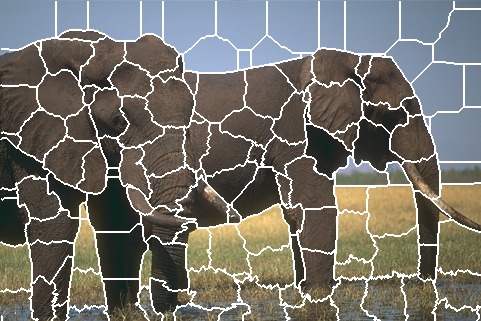}&
\includegraphics[width=\viscompwd]{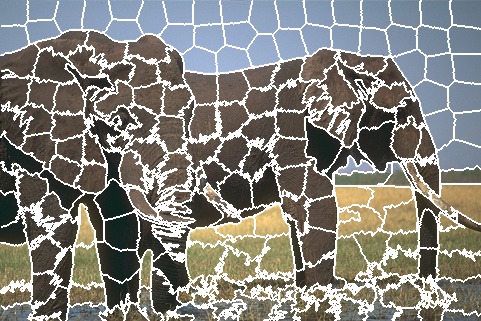}&
\includegraphics[width=\viscompwd]{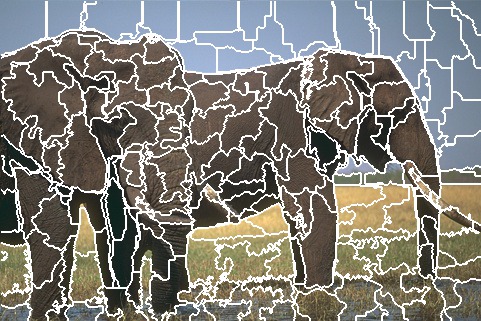}&
\includegraphics[width=\viscompwd]{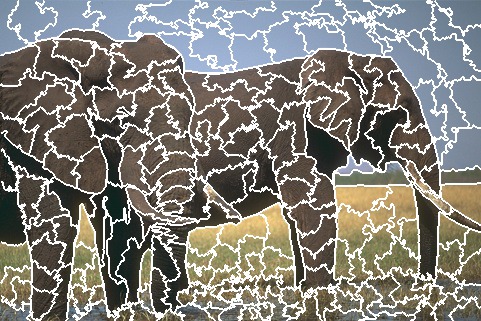}&
\includegraphics[width=\viscompwd]{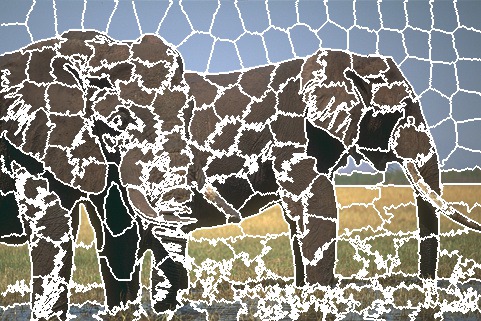}&
\includegraphics[width=\viscompwd]{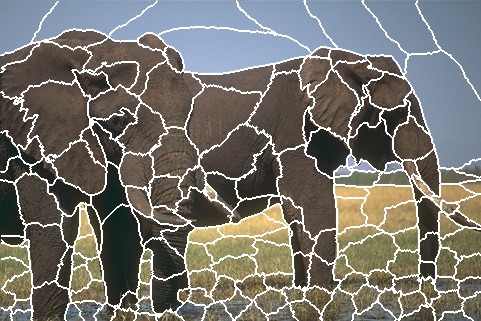}\\

\includegraphics[width=\viscompwd]{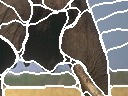}&
\includegraphics[width=\viscompwd]{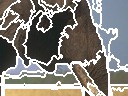}&
\includegraphics[width=\viscompwd]{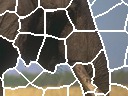}&
\includegraphics[width=\viscompwd]{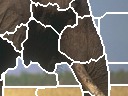}&
\includegraphics[width=\viscompwd]{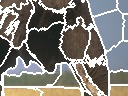}&
\includegraphics[width=\viscompwd]{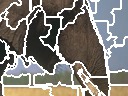}&
\includegraphics[width=\viscompwd]{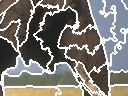}&
\includegraphics[width=\viscompwd]{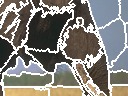}&
\includegraphics[width=\viscompwd]{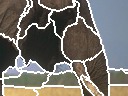}\\

\includegraphics[width=\viscompwd]{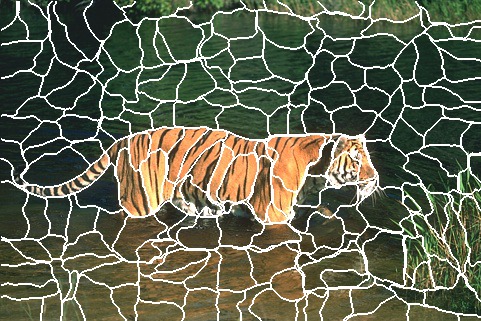}&
\includegraphics[width=\viscompwd]{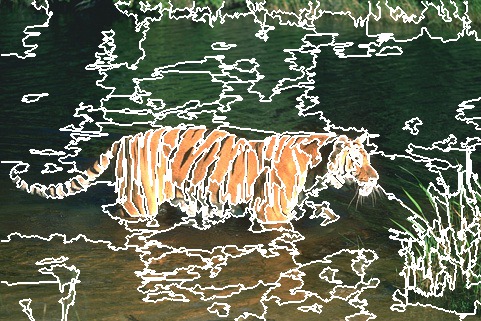}&
\includegraphics[width=\viscompwd]{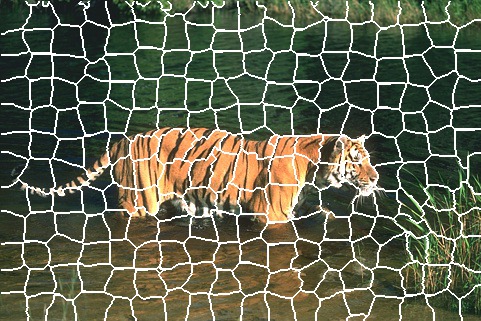}&
\includegraphics[width=\viscompwd]{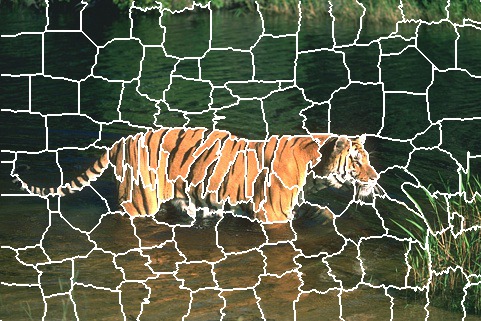}&
\includegraphics[width=\viscompwd]{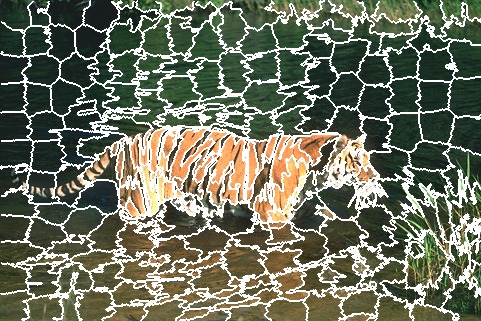}&
\includegraphics[width=\viscompwd]{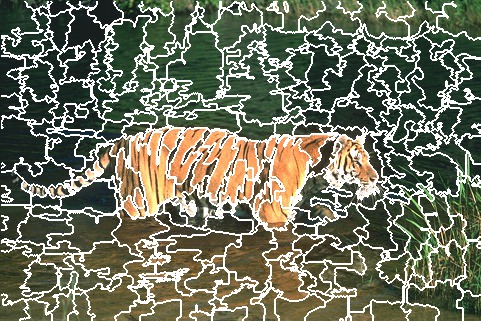}&
\includegraphics[width=\viscompwd]{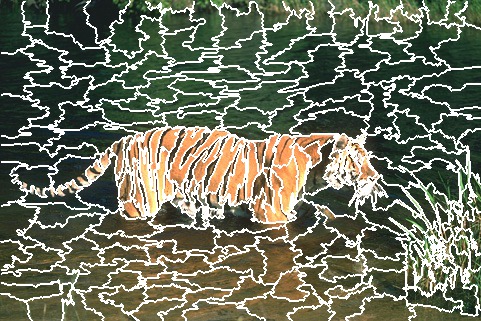}&
\includegraphics[width=\viscompwd]{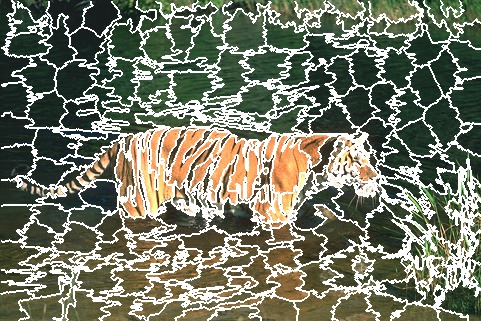}&
\includegraphics[width=\viscompwd]{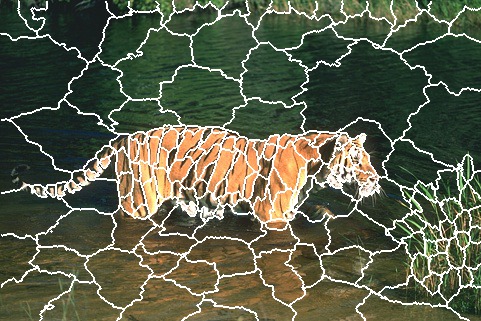}\\

\includegraphics[width=\viscompwd]{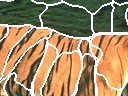}&
\includegraphics[width=\viscompwd]{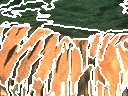}&
\includegraphics[width=\viscompwd]{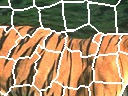}&
\includegraphics[width=\viscompwd]{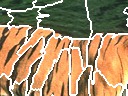}&
\includegraphics[width=\viscompwd]{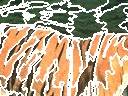}&
\includegraphics[width=\viscompwd]{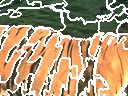}&
\includegraphics[width=\viscompwd]{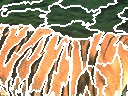}&
\includegraphics[width=\viscompwd]{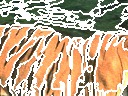}&
\includegraphics[width=\viscompwd]{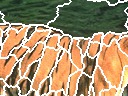}\\

{\small NCUTS~\cite{Shi_Malik_NCuts_PAMI2000}}& {\small
EGB~\cite{Felzenszwalb_Huttenlocher_IJCV2004}} & {\small TPIX~\cite{Levenshtein_Kutulakos_Fleet_Dickinson_Siddiqi_PAMI2009}} & {\small GCUT~\cite{Veksler_Superpixels_ECCV2010}} & {\small SLIC~\cite{Achanta_etal_SLIC_PAMI2012}} & {\small SEEDS~\cite{Bergh_etal_SEEDS_IJCV2015}} & {\small ERS~\cite{Liu_etal_ERS_CVPR2011}} & {\small LSC~\cite{Li_Chen_CVPR2015}} & {\small \textbf{Adaptels}}\\
\end{tabular}
\caption{Visual comparison of different segmentation algorithms. In the zoomed-in regions notice how well adaptels adhere to each region of the image according to its local structure. Results for adaptels use a threshold~$T=90\,\textrm{bits}$ and the initial
seed is the central pixel of each image.}
\label{fig:viscomp2}
\end{figure*}


\section{Comparison}
\label{sec:comparison}

We assess the power of the information uniformity assumption
against the standard size uniformity assumption. We compare adaptels to several state-of-the-art
superpixel methods: EGB~\cite{Felzenszwalb_Huttenlocher_IJCV2004}, TPIX~\cite{Levenshtein_Kutulakos_Fleet_Dickinson_Siddiqi_PAMI2009}, GCUT~\cite{Veksler_Superpixels_ECCV2010}
SLIC~\cite{Achanta_etal_SLIC_PAMI2012}, SEEDS~\cite{Bergh_etal_SEEDS_IJCV2015},
ERS~\cite{Liu_etal_ERS_CVPR2011} and LSC~\cite{Li_Chen_CVPR2015}. We used implementations available online for all methods~\cite{code_egb,code_slic,code_lsc,code_ers,code_gcuts,code_tpix,code_seeds}.

We use the Berkeley~300 dataset~\cite{Martin_etal_BerekelyDataset_ICCV2001} with its color and gray
scale groundtruth images (3269~in total).
Fig.~\ref{fig:plots} depicts results for the entire range of
50 to 2000~superpixels, corresponding to an image
simplification ranging from four to two orders of magnitude.

\subsection{Under-segmentation error}
Under-segmentation error measures the overlap error, also termed ``leak" or ``bleeding" between groundtruth and superpixel segments. The computation of under-segmentation error as presented in TPIX and later in SLIC penalizes every overlapping error twice, on either side of the erring superpixel as pointed out by Neubert and Protzel~\cite{Neubert_Protzel_2012}. We compute the Corrected Under-Segmentation Error (CUSE)~\cite{Neubert_Protzel_2012}. CUSE for each image is computed as the sum of overlap error for each superpixel segment $S_k$:

\begin{equation}
\label{eq:cuse}
{CUSE} = \frac{1}{N}\sum^{K}_{k=1}|G^{\ast}(S_k)\cup S_k - G^{\ast}(S_k)|
\end{equation}
\noindent where $G^{\ast}(S_k) = \argmax_{j}{(G_j\cap S_k)}$ is the ground truth segment with which segment $S_k$ has the maximum overlap.

A related comparison measure introduced by ERS~\cite{Liu_etal_ERS_CVPR2011}, and also computed by SEEDS~\cite{Bergh_etal_SEEDS_IJCV2015}, is Achievable Segmentation Accuracy (ASA), expressed as: $
{ASA} = \frac{1}{N}\sum^{K}_{k=1}G^{\ast}(S_k)$. We see that ASA is simply the complement of CUSE - for any given pair of superpixel and ground truth labels, the two values add up to $1$. We therefore only show the plot for CUSE in Fig.~\ref{fig:plots}. Adaptels show the least error of all for most superpixel sizes.

\subsection{Boundary recall}
Recall is the ratio of the true positives ($TP$) to the sum of true positives and false negatives ($FN$). We represent boundary maps, which have the same size and dimensions as the corresponding image, for superpixel segmentation as $b^S_i$, and for ground truth as $b^G_i$, such that the value at pixel position $i$ is $1$ in the presence of a boundary and $0$ otherwise. Boundary recall is computed for each pair of input image and groundtruth in the same way as done by TPIX~\cite{Levenshtein_Kutulakos_Fleet_Dickinson_Siddiqi_PAMI2009}, SLIC~\cite{Achanta_etal_SLIC_PAMI2012}, SEEDS~\cite{Bergh_etal_SEEDS_IJCV2015}, and ERS~\cite{Liu_etal_ERS_CVPR2011}:
\begin{equation}
\label{eq:recall}
{Recall} = \frac{TP}{TP+FN} = \frac{\sum^N_{i=1}{\mathds{I}_{j\in \mathcal{N}(i,\epsilon)}(b^G_i \land b^S_j)}}{\sum^N_{i=1}{\mathds{I}(b^G_i)}}
\end{equation}
\noindent where $\land$ represents a logical \emph{and} operation, $\mathds{I}$ represents a function that returns $1$ if the entity passed to the function is greater than $0$, and $\mathcal{N}(i,\epsilon)$ is the neighborhood of $i$ at range $\epsilon$. The denominator term $TP+FN$ is then simply the number of all boundary pixels. We use $\epsilon = 2$ as done in the past~\cite{Achanta_etal_SLIC_PAMI2012,Liu_etal_ERS_CVPR2011,Bergh_etal_SEEDS_IJCV2015,Neubert_Protzel_2012}.

\subsection{Boundary precision}
By treating a segmentation algorithm as a boundary detection algorithm, superpixel algorithms compute boundary recall for comparison. But recall alone can be misleading since it is possible to have a very high recall with extremely poor precision. For the task of segmentation, it is well established in literature~\cite{Martin_Fowlkes_Malik_PAMI2004,Arbelaez_etal_PAMI2011} that recall has to be regarded in conjunction with precision.

In this paper, we compute precision, which is often missing in previous works (TP~\cite{Levenshtein_Kutulakos_Fleet_Dickinson_Siddiqi_PAMI2009}, SLIC~\cite{Achanta_etal_SLIC_PAMI2012}, SEEDS~\cite{Bergh_etal_SEEDS_IJCV2015}, ERS~\cite{Liu_etal_ERS_CVPR2011}). To compute precision, we need to know the number of false positives $FP$, which is the number of superpixel boundary pixels in the $\epsilon$ neighbourhood that are not true positives:
\begin{equation}
\label{eq:precision}
FP = \sum^N_{i=1}{\left[1 - \mathds{I}_{j\in \mathcal{N}(i,\epsilon)}(b^G_i \land b^S_j)\right]}
\end{equation}
\noindent Knowing $FN$ allows us to compute precision as ${Precision} = TP/(TP+FP)$ where $TP$ is the same as the numerator term of Eq.~\ref{eq:recall}.

\subsection{Precision-recall and F-measure}
\begin{figure}[h]
\centering
\begin{tabular}{c}
\includegraphics[width=1.0\linewidth]{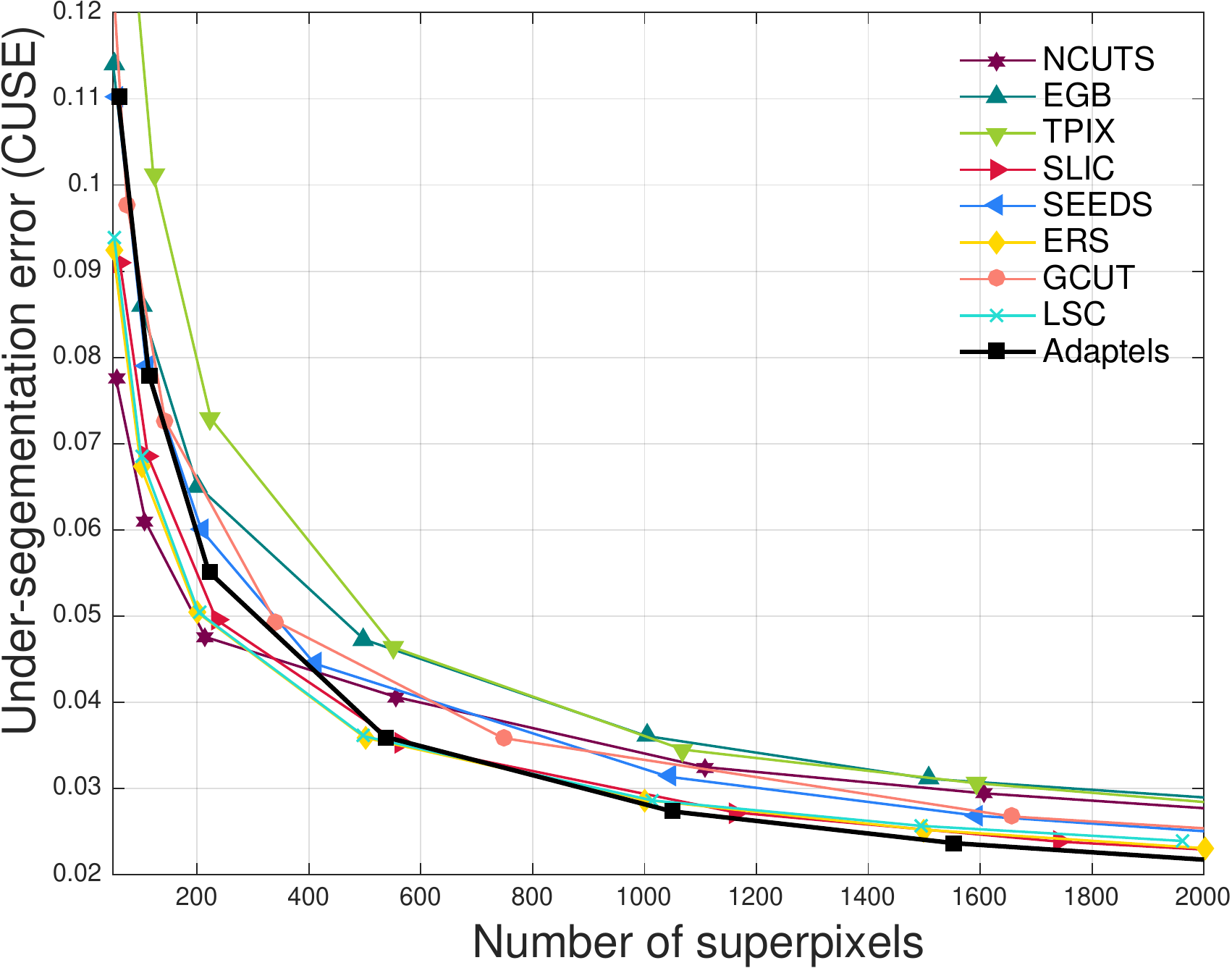}\\
\includegraphics[width=1.0\linewidth]{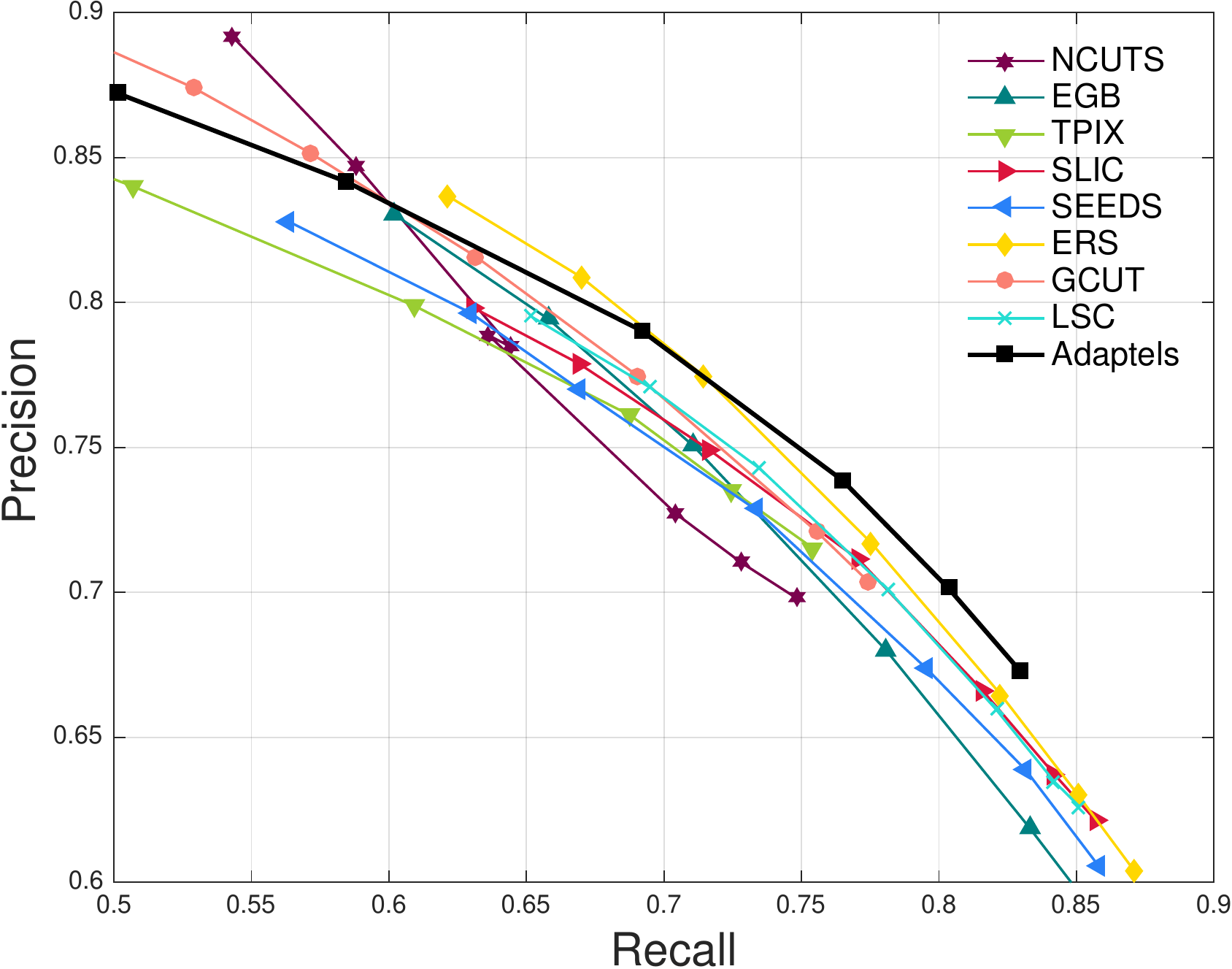} \\
\includegraphics[width=1.0\linewidth]{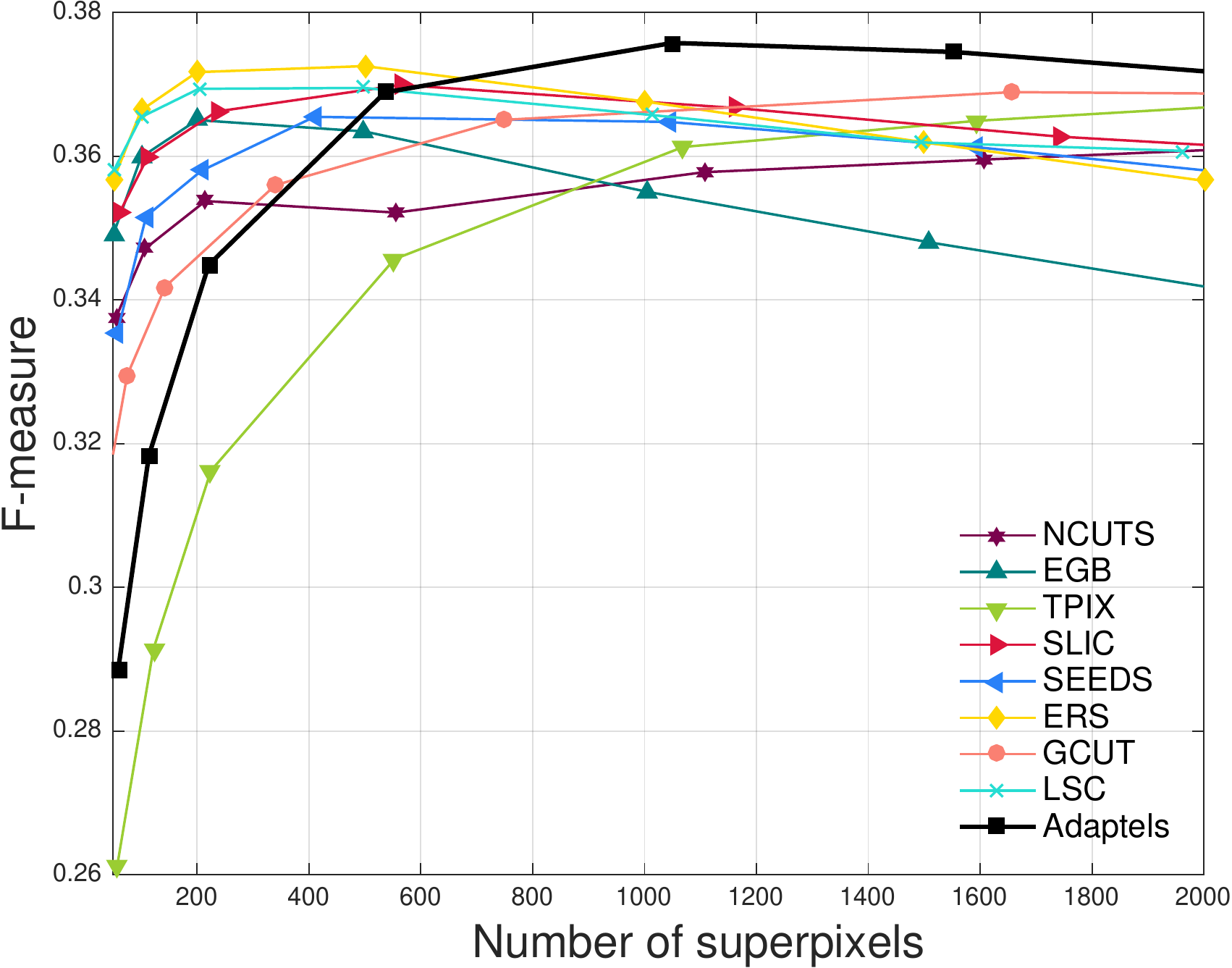} \\
\end{tabular}

\caption{(Top)~The under-segmentation error (CUSE) of adaptels is the lowest. (Middle)~Adaptels have a better precision-versus recall performance, and (Bottom) a better F-measure than the state-of-the-art. The values at 1000 superpixels are compared in Table~\ref{table:numerical_comparison}.}
\label{fig:plots}
\end{figure}

Using the boundary recall and precision values, we are able to plot the more conclusive curves of Precision vs. Recall and F-measure versus number of superpixels, shown in Fig.~\ref{fig:plots}. These plots prove the superiority of adaptels over other methods.

\begin{table}
\centering
\begin{tabular}{l r r r}
\\
\hline
\small
      & CUSE & F-measure & Speed (fps) \\
\hline
\hline
NCUTS & 0.0325 & 0.3578 & -\\
EGB & 0.0361 & 0.3551 & 12.5\\
TPIX & 0.0345 & 0.3613 & 0.2\\
GCUTS & 0.0358 & 0.3651 & 0.5\\
SLIC & 0.0274 & 0.3667 & 13.9\\
SEEDS & 0.0313 & 0.3648 & \textbf{18.1}\\
ERS & 0.0286 & 0.3676 & 1.2\\
LSC & 0.0286 & 0.3657 & 3.2\\
\textbf{Adaptels} & \textbf{0.0273} & \textbf{0.3758} & 14.7\\
\hline
\\
\end{tabular}
\caption{Every row shows the values computed for a given algorithm for the corresponding metric from Fig.~\ref{fig:plots} for 1000~superpixels on images of size $321 \times 481$ pixels. The best values are highlighted in bold. Adaptels outperform the other methods on these metrics. NCUTS is far slower than other methods, taking over a hundred seconds for a single image. Apart from SEEDS, whose timing varies from 12 to 24~fps depending on the number of superpixels, the Adaptel algorithm is the fastest method, capable of processing nearly 15~fps on a regular laptop.}
\label{table:numerical_comparison}
\end{table}

\subsection{Computational efficiency}
We measured the speed of every method for
images of size $481 \times 321$. We present the average number of frames-per-second (fps)
in Table~\ref{table:numerical_comparison}. All of the algorithms run
on the same hardware (2.6 GHz Intel Core i7 processor, with 16 GB of RAM, running OSX).
We do not use any parallelization, GPU processing, or dedicated
hardware for any of the algorithms. The speed of SEEDS varies from 12 to 24~fps
for different number of superpixels. The speed of other algorithms, including Adaptels,
is independent of the number of superpixels. Barring SEEDS, the Adaptel is the fastest segmentation algorithm.

\subsection{Discussion of results}
Adaptels exhibit the lowest under-segmentation error of all methods compared with (Fig.~\ref{fig:plots}).  As is customary for any detection problem, we compute boundary precision along with boundary recall. The two measures independently are insufficient to convey the quality of a segmentation algorithm. For instance, it is possible to have a very high recall if all superpixels are of size $1$. Similarly, it is possible to have high precision even if a single boundary pixel is correctly detected and there are no false positives. Considering the two values together avoids biasing the evaluations towards methods that generate noisy or jagged segment boundaries, e.g. EGB, SLIC, ERS, and LSC.

With the help of these two measures we plot the precision-recall curve and F-measure curve shown in Fig.~\ref{fig:plots}. These curves offer a fair and conclusive comparison of the segmentation methods. In both these plots we notice that adaptels outperform the state-of-the-art. Additionally, the numbers provided in Table~\ref{table:numerical_comparison} show that adaptels not only outperform the state-of-the-art in terms of segmentation quality but also in terms of computational efficiency.

\section{Video adaptels}
\label{sec:videoseg}

\begin{figure}[htbp]
\centering
\begin{tabular}{c}
\includegraphics[width=\linewidth]{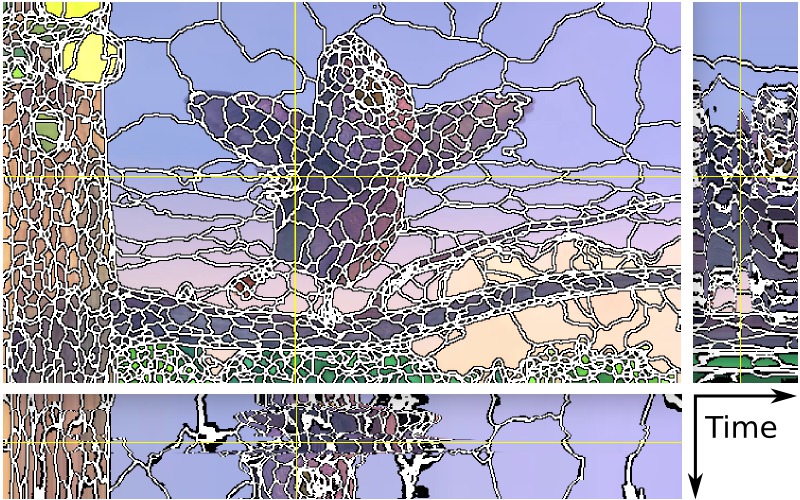}\\

\end{tabular}
\caption{Adaptel segmention in 3D. This image shows a cross-hairs based 3D section of the segmented video volume of 100 frames seen from the right (Y-Time plane) and bottom (X-Time plane). The segmented video is provided in the supplementary material.}
\label{fig:supervoxels}
\end{figure}

It is trivial to extend the Adaptel algorithm to higher dimensional data like image stacks and video volumes. Adaptel segmentation starts at the center of the volume. Instead of looking in the 2D neighborhood for growing an Adaptel, connected pixels are obtained from a 3D neighborhood. So, apart from this change to $\mathcal{N}$ and the necessary changes to the maps $C$ and $D$, the other steps remain the same in algorithms~\ref{algo:adaptel} and \ref{algo:adaptels} for 3D segmentation.  Fig.~\ref{fig:supervoxels} shows an example of a segmented video volume with the cross-sections along the time axis. Fig.~\ref{fig:voxels} shows a few individual frames from the same video volume. Just as in 2D, the object boundaries are also well-adhered to in the 3D case. The computational complexity remains linear in the number of voxels in the volume.

\newcommand{\vidframecolwd}{0.3\linewidth}
\newcommand{\vidframecolwdvox}{0.5\linewidth}
\newcommand{\voxgap}{@{\hspace{0.5mm}}}

\begin{figure}
\centering
\begin{tabular}{\gap c \gap c}
\includegraphics[width=\vidframecolwdvox]{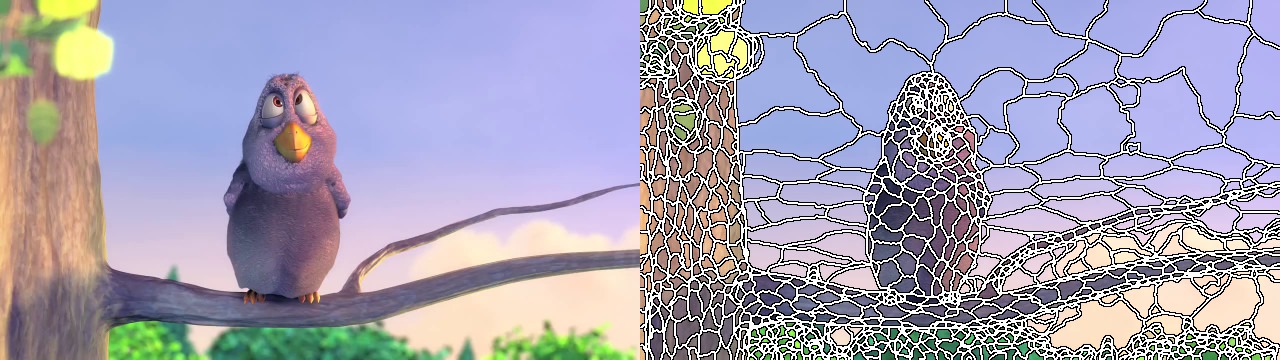}&
\includegraphics[width=\vidframecolwdvox]{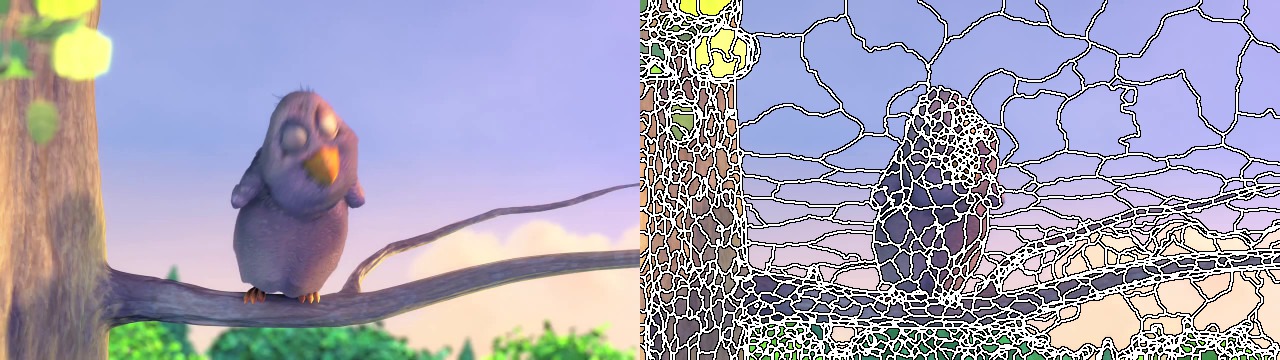}\\
\includegraphics[width=\vidframecolwdvox]{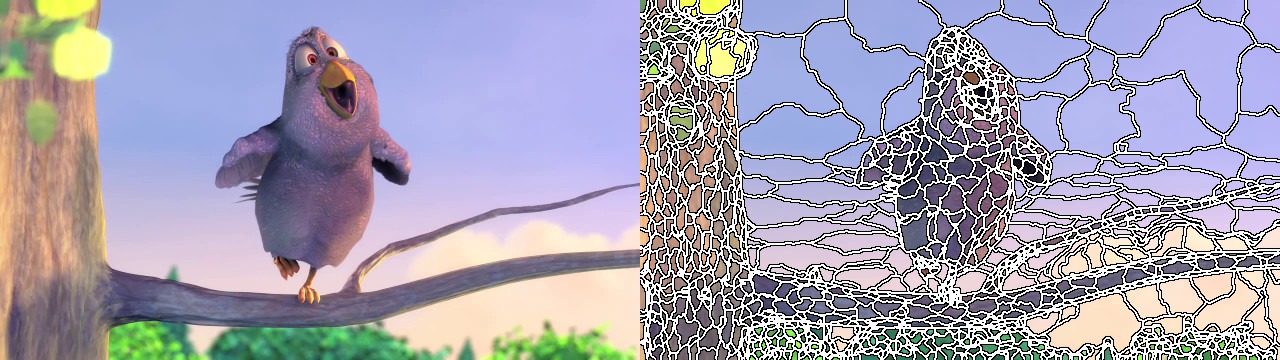}&
\includegraphics[width=\vidframecolwdvox]{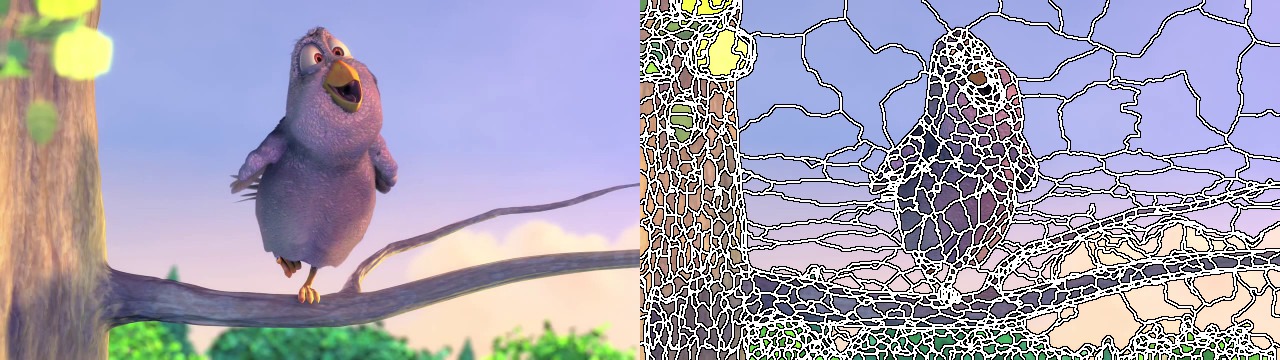}\\
\includegraphics[width=\vidframecolwdvox]{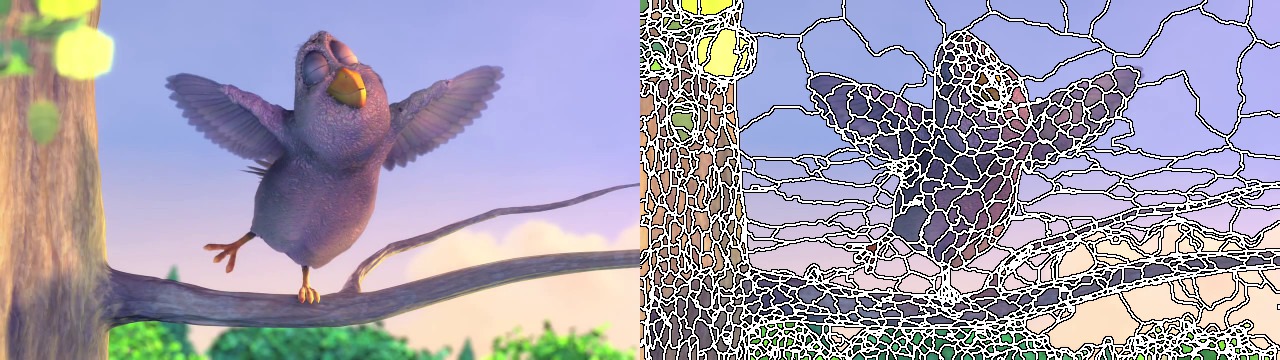}&
\includegraphics[width=\vidframecolwdvox]{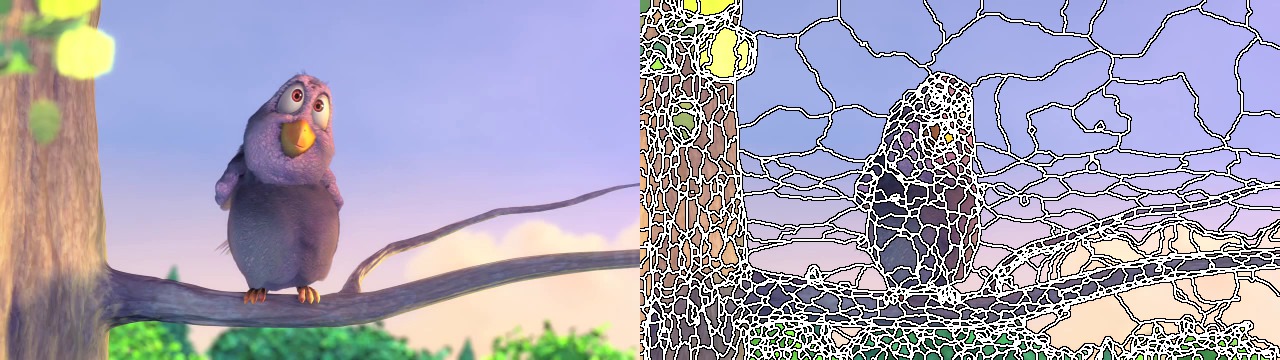}\\
\includegraphics[width=\vidframecolwdvox]{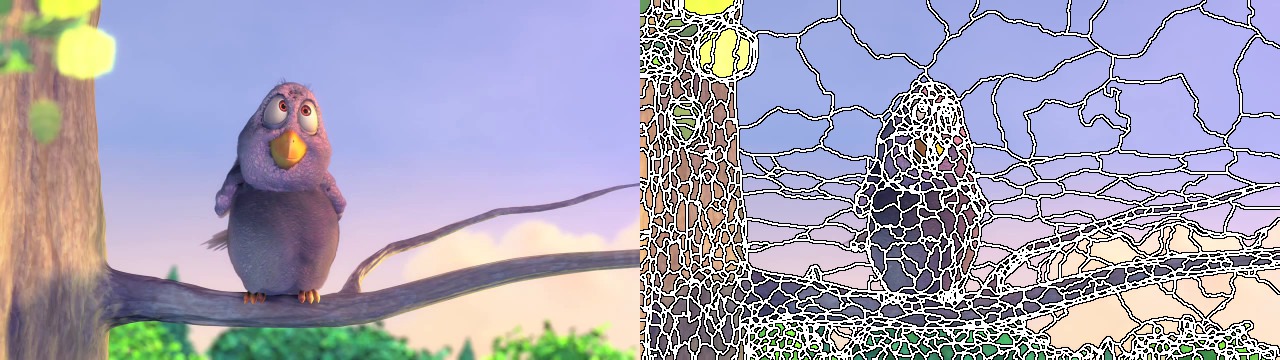}&
\includegraphics[width=\vidframecolwdvox]{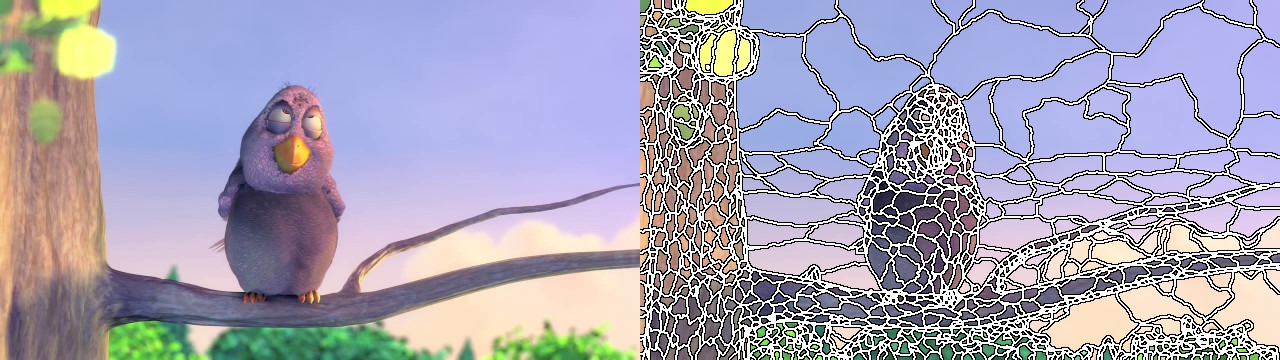}\\
\includegraphics[width=\vidframecolwdvox]{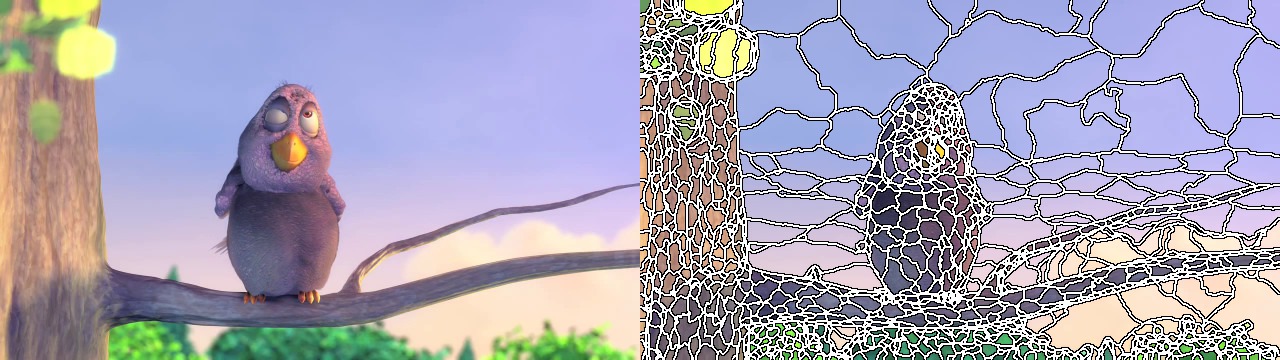}&
\includegraphics[width=\vidframecolwdvox]{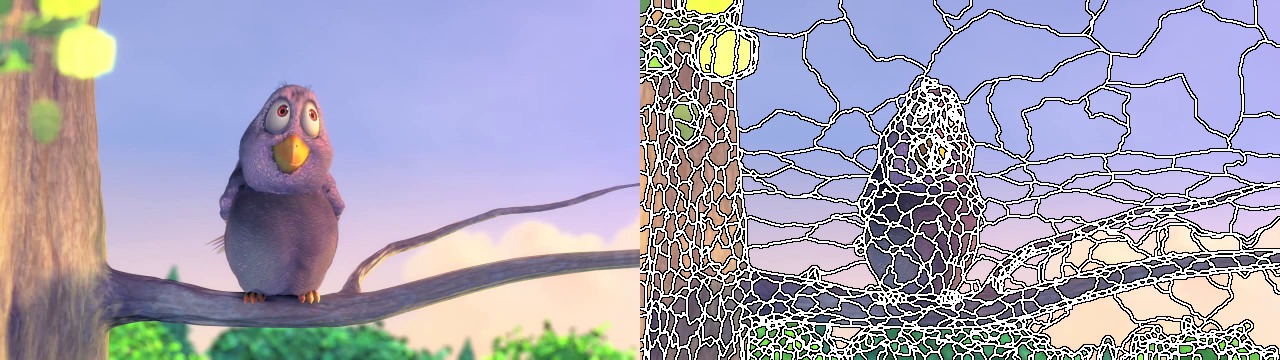}\\
\end{tabular}
\caption{Supervoxel segmentation. The images here show a few frames of the video volume of Fig.~\ref{fig:supervoxels} that has been segmented using the 3D extension of the Adaptel algorithm.}
\label{fig:voxels}
\end{figure}

\section{Conclusion}
\label{sec:conclusion}
We introduced an information-theoretic approach for creating
image segments which we call \emph{Adaptels}. Instead of assuming uniformity in
size or shape, we assume information uniformity. This leads to
segments that change their size according to the local image
complexity as defined by a family of probability distributions.
We have used the double exponential distribution to encode complexity in natural images. For other types of images, our approach offers the possibility of using application-specific distributions.


The experimental comparison proves the superiority of adaptels.
Our algorithm has the lowest under-segmentation error, is the best in both precision versus recall performance, as well as in terms of F-measure. The algorithm is linear in the number of pixels,
and is among the fastest methods considered. It is simple to use,
requiring only one parameter: the upper-bound of the information contained
in every segment. There is no need to set the superpixel size or choose seeds a priori, since both of these are achieved automatically by the algorithm. Finally, the algorithm is easily extended to higher dimensional data.

\end{document}